\documentclass[letterpaper]{article}

\usepackage{amsmath}
\usepackage{natbib,alifeconf}  
\usepackage{caption,subcaption}
\usepackage{hyperref}
\usepackage{etoolbox}

\usepackage[T1]{fontenc}
\newtoggle{blindsubmission}

\togglefalse{blindsubmission}

\title{Growing Steerable Neural Cellular Automata}

\iftoggle{blindsubmission}{
\author {Anonymous Authors}
}{
\author {Ettore Randazzo,$^{1}$ Alexander Mordvintsev,$^{1}$ \and Craig Fouts$^{2}$\\
\mbox{} \\
$^1$ Google Research \\
\{etr, moralex\}@google.com \\
\\
$^2$ Columbia University\\
cwf2117@columbia.edu}
}

\begin{document}
\maketitle

\begin{abstract}

Neural Cellular Automata (NCA) models have shown remarkable capacity for pattern formation and complex global behaviors stemming from local coordination. However, in the original implementation of NCA, cells are incapable of adjusting their own orientation, and it is the responsibility of the model designer to orient them externally. A recent isotropic variant of NCA (Growing Isotropic Neural Cellular Automata) makes the model orientation-independent - cells can no longer tell up from down, nor left from right - by removing its dependency on perceiving the gradient of spatial states in its neighborhood. In this work, we revisit NCA with a different approach: we make each cell responsible for its own orientation by allowing it to "turn" as determined by an adjustable internal state. The resulting Steerable NCA contains cells of varying orientation embedded in the same pattern. We observe how, while Isotropic NCA are orientation-agnostic, Steerable NCA have chirality: they have a predetermined left-right symmetry. We therefore show that we can train Steerable NCA in similar but simpler ways than their Isotropic counterpart by \textbf{(1)} breaking symmetries using only two seeds, or \textbf{(2)} introducing a rotation-invariant training objective and relying on asynchronous cell updates to break the up-down symmetry of the system.

\end{abstract}

\section{Introduction}

Recently, Neural Cellular Automata (NCA) have shown remarkable capacity for discovering complex rules for a variety of tasks, including digit classification \citep{randazzo2020self-classifying, walker2022physical}, texture generation \citep{niklasson2021self-organising,mordvintsev2021differentiable}, morphogenesis \citep{mordvintsev2020growing, mordvintsev2022iso, sudhakaran2021growing, NEURIPS2021_af87f7cd}, adversarial attacks \citep{randazzo2021adversarial, cavuoti2022}, and control systems \citep{alex2021selforganized}. For the most part, implementations of such NCA are \textit{anisotropic}: cells are directionally-dependent on some external factors, such as a predetermined orientation and coordinate system or an oriented perception kernel.
Anisotropy may be an undesirable property, as it makes it impossible to produce groups of cells with inherently different orientations unless externally defined, and it relies on non-local information (such as a grid- or axis-alignment) that may or may not be available in practical applications.

\begin{figure*}[ht]
  \includegraphics[width=\textwidth]{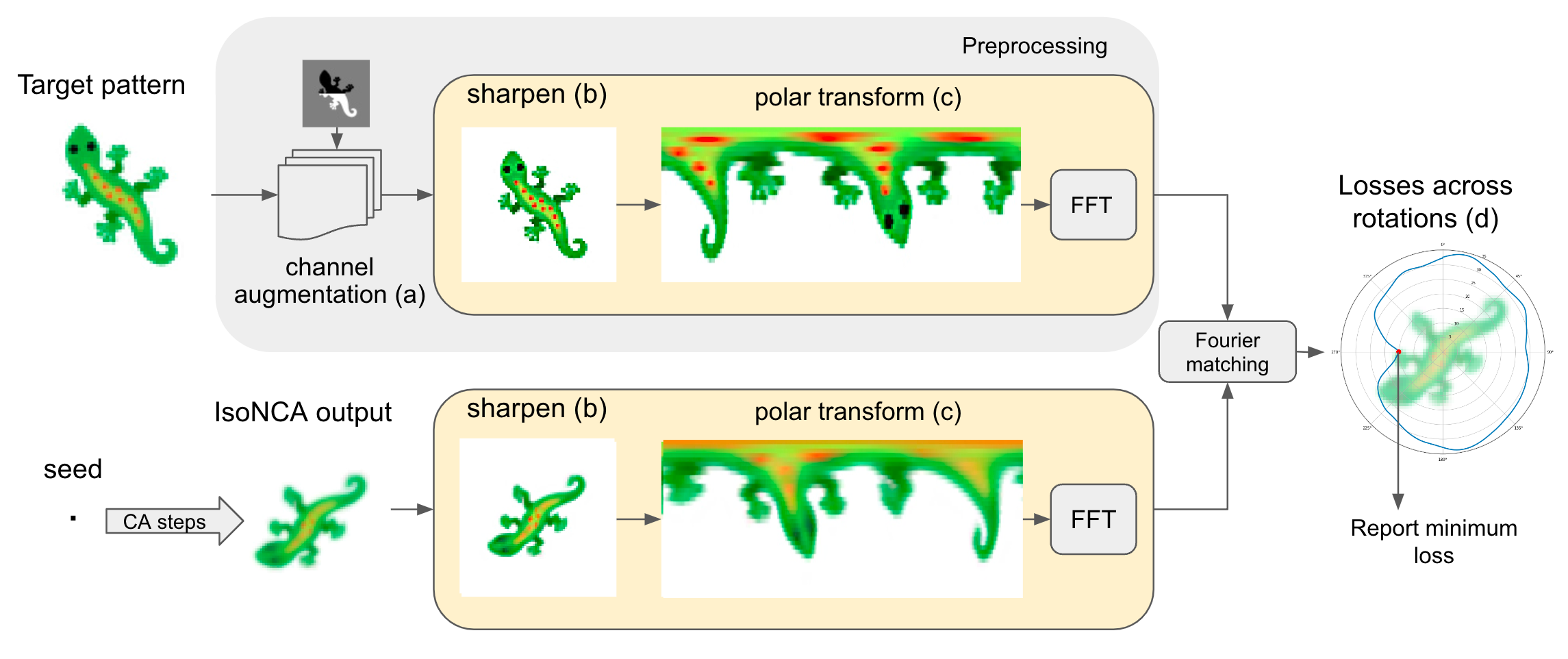}
  \caption{Rotation-invariant Steerable NCA training pipeline. The target pattern is augmented with extra channels \textbf{(a)} to break symmetries that may interfere with training. Both the target- and NCA-grown patterns are sharpened \textbf{(b)} to steer optimization into preserving fine details. A polar transformation \textbf{(c)} is applied to turn the unknown rotation between two images into a horizontal shift. Fourier-domain image matching then enables efficient computation of pixel-matching losses across all orientations \textbf{(d)}. The blue plot shows losses with respect to all rotations of the original pattern. The minimal loss value is selected for backpropagation.}
  \label{fig:invariant_loss}
\end{figure*}

\subsection{Isotropy in Cellular Automata}

Isotropic cellular automata is not a novel concept: indeed, Conway's Game of Life is isotropic \citep{10.2307/24927642}. Generalizations, such as Lenia \citep{chan2019lenia}, are also inherently isotropic and rely on the initial configuration of cells for the emergence of complex behaviours. Until very recently, however, discovering complex isotropic rules has been achieved with limited success, with procedures mostly dependent on manual search and evolutionary strategies. 

With the advent of NCA, end-to-end differentiable optimization has provided a powerful medium for discovering more complex rules. An example of fully isotropic NCA is presented in \cite{mordvintsev2021differentiable}, where the NCA can be seen as a reaction-diffusion system. However, the loss used is a rotation-invariant variation of the texture loss used in \cite{niklasson2021self-organising}, which is not invariant to reflections, and the resulting textures are not capable of creating very complex shapes.
In a more recent work \citep{mordvintsev2022iso}, a fully isotropic NCA (IsoNCA) is introduced for complex pattern formation. This is the first study on NCA that explicitly addresses issues with training fully isotropic pattern growth, such as the inherent rotation \textit{and} reflection symmetry of any target pattern. The authors solved these issues with two alternative approaches: (1) by breaking the up-down and left-right symmetries of the system with the initial conditions by means of \textit{structured seeds}, and (2) by introducing a novel rotation-reflection invariant loss with additional enhancements to facilitate training.

We also want to highlight that, since NCA can be seen as a special case of Graph Neural Networks, one can construct them to be isotropic \citep{NEURIPS2021_af87f7cd, gala2023enequivariant}. The focus of this paper, however, is on the more traditional grid-based NCA, such as in IsoNCA, where cells are immovable and their perception is restricted by a local neighborhood that cannot be arbitrarily distant.

\subsection{Chirality and Steerable NCA}

While IsoNCA might be an appropriate choice of architecture for a variety of tasks, sometimes the property of \textit{reflection-invariance} is unnecessary or even undesirable. Many biological molecules and structures, such as proteins, sugars, nucleic acids, and amino acids, exhibit chirality. It is possible that chirality at this level helps break asymmetry at higher levels such as organs, which are highly asymmetric in most animals \citep{Inaki2016-uz}.

Inspired by biological cells, we develop and train NCA that have an inherent cellular chirality, which we call \textit{Steerable NCA} due to their capacity to self-orient their perception field. Steerable NCA are capable of perceiving gradients of states in their neighborhood, like in \cite{mordvintsev2020growing}, but the orientation of the perception field is dependent on internal states which may vary from cell to cell. Since the orientation is exclusively defined by an angle for each cell, the x-y gradients have an inherent chirality.

Similarly to IsoNCA, we show that Steerable NCA can be trained in two possible ways: (1) Steerable NCA can be initialized with two different seeds - as opposed to the minimum of 3 required to train IsoNCA - positioned near one another to break the up-down symmetry (where 'seeds' refers to alive cells placed in the initial state of the system), and (2) a single seed can be used alongside a \textit{rotation-invariant loss} - as opposed to the rotation-reflection variant used for IsoNCA - and stochastic updates are responsible for breaking the up-down symmetry of the system. These simplifications are possible because Steerable NCA need only break a single up-down symmetry, as opposed to both up-down \textit{and} left-right required by IsoNCA, resulting in simpler and faster training regimes.

\iftoggle{blindsubmission}{
We plan to release a set of Google Colabs to easily reproduce the entirety of the experiments presented in this paper.
}{
We released a set of Google Colab notebooks to easily reproduce the entirety of the experiments presented in this paper, available at https://github.com/google-research/self-organising-systems/tree/master/isotropic\_nca.
}

\section{Steerable Neural CA model}

The Steerable NCA model described here can be seen as a variant of the Growing NCA model \citep{mordvintsev2020growing}, differing only in its perception phase. This section covers key features of the model design. Below, we adapt the equivalent section in \citep{mordvintsev2022iso} to better showcase the similarities and differences between the two approaches.

\paragraph{Grid} Cells exist on a regular Cartesian grid; the state of each cell is represented by a vector $$\mathbf{s} = [s^0=R, s^1=G, s^2=B, s^3=A, ... , s^{C-1}]  \label{eq:s}$$ where $C$ is the number of channels and the first four channels represent a visible RGBA image. The whole grid is initialized with zero values, except for the seed cell in which $A=1$. Setting the $A$ value to 1 ensures that the seed starts in an alive configuration, as we will explain later. Cells then iteratively update their states using only the information collected from their 3x3 Moore neighbourhood.

\paragraph{Stochastic updates} Cell updates occur stochastically: at each NCA step, cells are updated with probability $p_\text{upd}$ (we use $p_\text{upd}=0.5$ in our experiments). This stochasticity is meant to eliminate dependence on a global shared clock that synchronizes the updates between cells. Previous work on NCA discusses the impact of this strategy on NCA robustness \citep{niklasson2021asynchronicity}. In Isotropic and Steerable NCA models, asynchronicity plays a critical role in the symmetry-breaking process; this asynchronicity can be seen as a strategy for generating noise, which has been documented to help biological systems construct complex functions in simple ways \citep{Samoilov2006-xj}.

\paragraph{``Alive'' and ``empty'' cells} The alpha channel ($s^3=A$) plays a special role in determining whether a cell is currently ``alive'' or ``empty''; here, a cell is alive if $A>0.1$ or if it has at least one alive cell in its 3x3 neighbourhood. The state of empty cells is explicitly set to zeros after each NCA step.

\paragraph{Perception} Each cell collects information about the state of its neighborhood using three filters: a per-channel discrete 3x3 Laplacian filter $K_{lap}$, and Sobel filters $K_{x}, K_{y}$ for the x- and y-axis respectively. The Laplacian filter computes the difference between the state of the cell and the average state of its neighbours, where:
$$
K_{lap} = 
\begin{bmatrix}
1 & 2 & 1\\2 & -12 & 2 \\1 & 2 & 1 \\
\end{bmatrix},\quad
K_{x} = 
\begin{bmatrix}
-1 & 0 & 1\\-2 & 0 & 2 \\-1 & 0 & 1 \\
\end{bmatrix}
$$
and $K_{y} = K_{x}^\top$.

The Sobel filters accumulate information about state gradients along their respective axis, resulting, for each cell, in a gradient vector for the x-axis, $g_x$, and one for the y-axis, $g_y$. In Growing NCA, these state gradients were rotated by a predetermined angle, $\theta$, to perceive correspondingly rotated x- and y-axis gradients. In Steerable NCA, each cell determines its own rotation angle, which it uses to rotate its perception field accordingly. In this work, we show two variants of Steerable NCA that accomplish the same goal in different ways: \textit{angle-} and \textit{gradient-}based rotations.

We highlight that the addition of the Laplacian filter (missing in the original Growing NCA paper) greatly improves training speed and performance with respect to models whose perception is solely made up of Sobel filters.

\paragraph{Angle-based rotation}
In this variant, each cell maintains its orientation angle, $\mathbf{s}_\theta$, as one of its state channels. The cell cannot perceive this angle in any way, but the update rule can still modify it, causing the cell to rotate its perception over time.
For each step, the final perception state $\mathbf{p}$ is computed as follows, with input $\mathbf{x} = (\mathbf{s}, \mathbf{s}_\theta)$:
\begin{align*}
g_x\, &= K_{x} \ast \mathbf{s},\quad g_y = K_{y} \ast \mathbf{s} \\
\mathbf{p}_x &= g_x cos(\mathbf{s}_\theta) + g_y sin(\mathbf{s}_\theta) \\
\mathbf{p}_y &= g_y cos(\mathbf{s}_\theta) - g_x sin(\mathbf{s}_\theta) \\
\mathbf{p}\;\: &= concat(\mathbf{s},\,K_{lap} \ast \mathbf{s},\,\mathbf{p}_x,\,\mathbf{p}_y)
\end{align*}

This variant is very intuitive, but it requires attention - rotating the cells' angle states accordingly - while performing external rotations on groups of cells to prevent them from behaving unexpectedly. The gradient-based variant is designed to circumvent this issue.

\paragraph{Gradient-based rotation}
In this variant, the orientation angle for each cell is dynamically inferred from the cell's respective neighborhood. To accomplish this, a concentration state, $\mathbf{s}_{conc}$, is passed through the Sobel filters to determine its respective gradients in the x- and y-direction. This pair of numbers is then normalized, and the resulting x and y normalized gradients are respectively used as the cosine and sine of the cell's implicit rotation angle. To stabilize computations, we clip the minimum value of the dividend to 1 during normalization. This has the effect that when the gradient of $\mathbf{s}_{conc}$ is minimal, the perceived state gradients are effectively multiplied by zero. For each step, the final perception state $\mathbf{p}$ is computed as follows, with input $\mathbf{x} = (\mathbf{s}, \mathbf{s}_{conc})$:
\begin{align*}
g_{conc} &= [K_{x} \ast \mathbf{s}_{conc},\, K_{y} \ast \mathbf{s}_{conc}] \\
[cos_\theta, sin_\theta] &= \frac{g_{conc}}{\textit{Clip}(\textit{Norm}(g_{conc}), 1, \infty)} \\
g_x\, &= K_{x} \ast \mathbf{s},\quad g_y = K_{y} \ast \mathbf{s} \\
\mathbf{p}_x &= g_x cos_\theta + g_y sin_\theta \\
\mathbf{p}_y &= g_y cos_\theta - g_x sin_\theta \\
\mathbf{p}\;\: &= concat(\mathbf{x},\, K_{lap} \ast \mathbf{x},\,\mathbf{p}_x,\,\mathbf{p}_y)
\end{align*}

Note how in this variant, cells can perceive $\mathbf{s}_{conc}$ without incurring any issues; as introduced in the previous paragraph, if one were to rotate a group of cells externally, the resulting system would not abruptly vary its behaviour.

\paragraph{Update rule} Cells stochastically update their states using a learned rule that is represented by a two-layer neural network: $$\mathbf{s}_{t+1} = \mathbf{s}_t + relu(\mathbf{p}_tW_0+b_0)W_1$$
where parameters $W_0$, $b_0$, and $W_1$ have shapes $(32, 192)$, $(192)$ and $(192, 16)$, respectively, giving a total of 9408 learned parameters.

\section{Training Steerable NCA}

As introduced before, Steerable NCA can be trained in two different ways: (1) by breaking symmetries during initialization using two \textit{different} seeds, or (2) by starting with a single seed and using a rotation-invariant loss function to train the system to break up-down symmetry itself.

\subsection{Two seeds strategy}

\begin{figure}[h]
 \centering
  \includegraphics[width=\columnwidth]{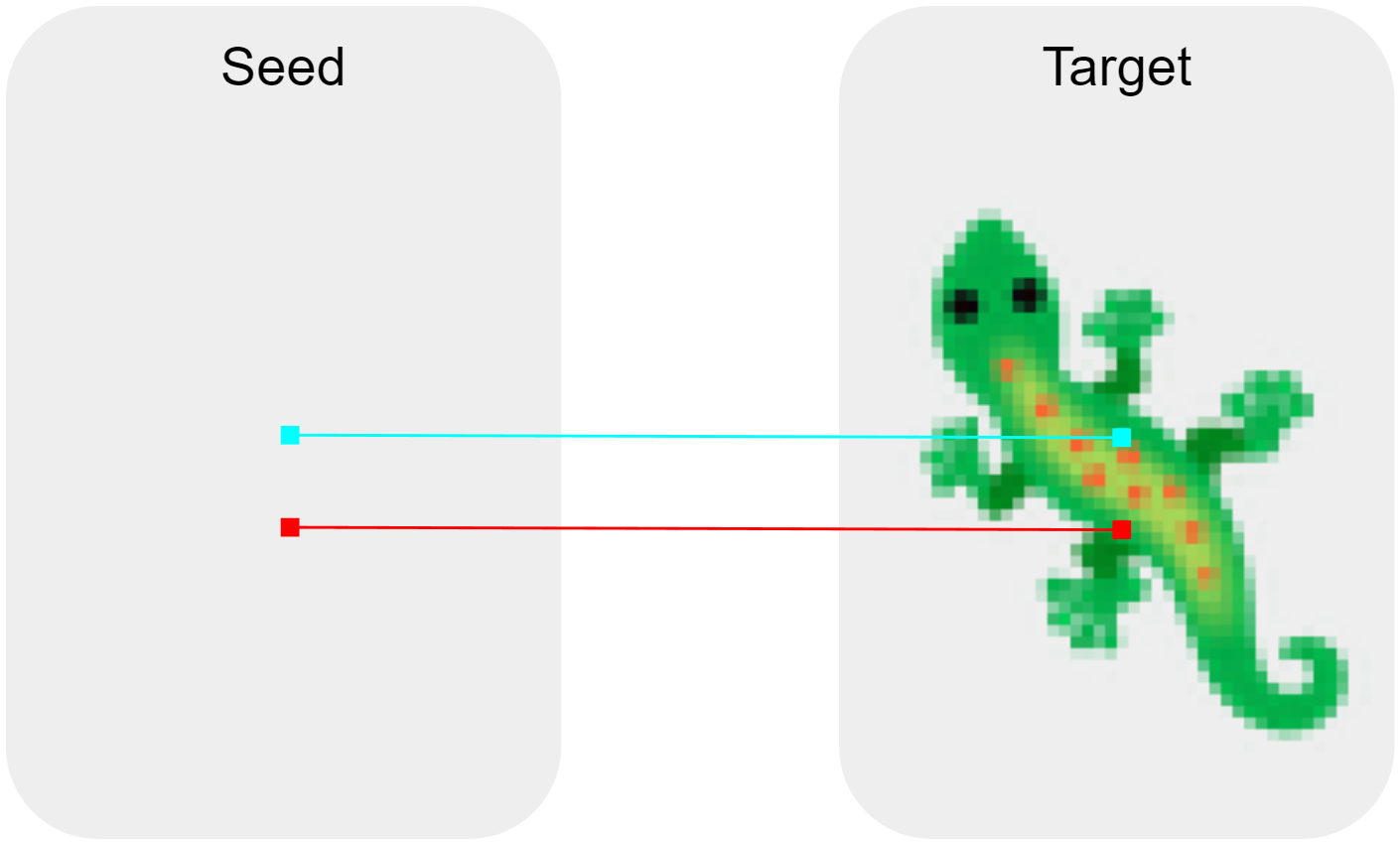}
  \caption{Two-seed mapping paradigm from which Steerable NCA can be trained to grow with the traditional pixel-wise L2 loss and without any additional auxiliary losses.}
  \label{fig:seed_map}
\end{figure}

By initializing the system with two different seeds - distinguished by their RGB configuration and placed slightly apart from one another - we define a more restrictive initial condition which implicitly resolves up-down symmetry. Doing so enables us to choose an arbitrary "training alignment" for the target pattern with which we can perform pixel-wise L2 loss, identically to the training procedure in Growing NCA. Here, just like IsoNCA, each seed serves as a point of origin for one of two sub-regions of the resultant shape. However, unlike IsoNCA, the chirality of our model precludes the necessity of left-right symmetry breaking, allowing us to train the model with two seeds instead of three (see Figure~\ref{fig:seed_map} for one possible two-seed mapping).

To generate visually- and spatially-differentiated seeds, we map their RGB encodings to distinct HSV hues and separate them by eight cells. While the exact specification is arbitrary, this procedure guarantees that (1) the seeds lie within the target pattern, and (2) they can be distinguished from one another by the model. As we'll see in the next section, this differentiation enables us to implicitly specify the model's orientation by imposing a rotation on the seed structure.

\subsection{Single seed strategy}

\begin{figure}[h]
 \centering
  \includegraphics[width=\columnwidth]{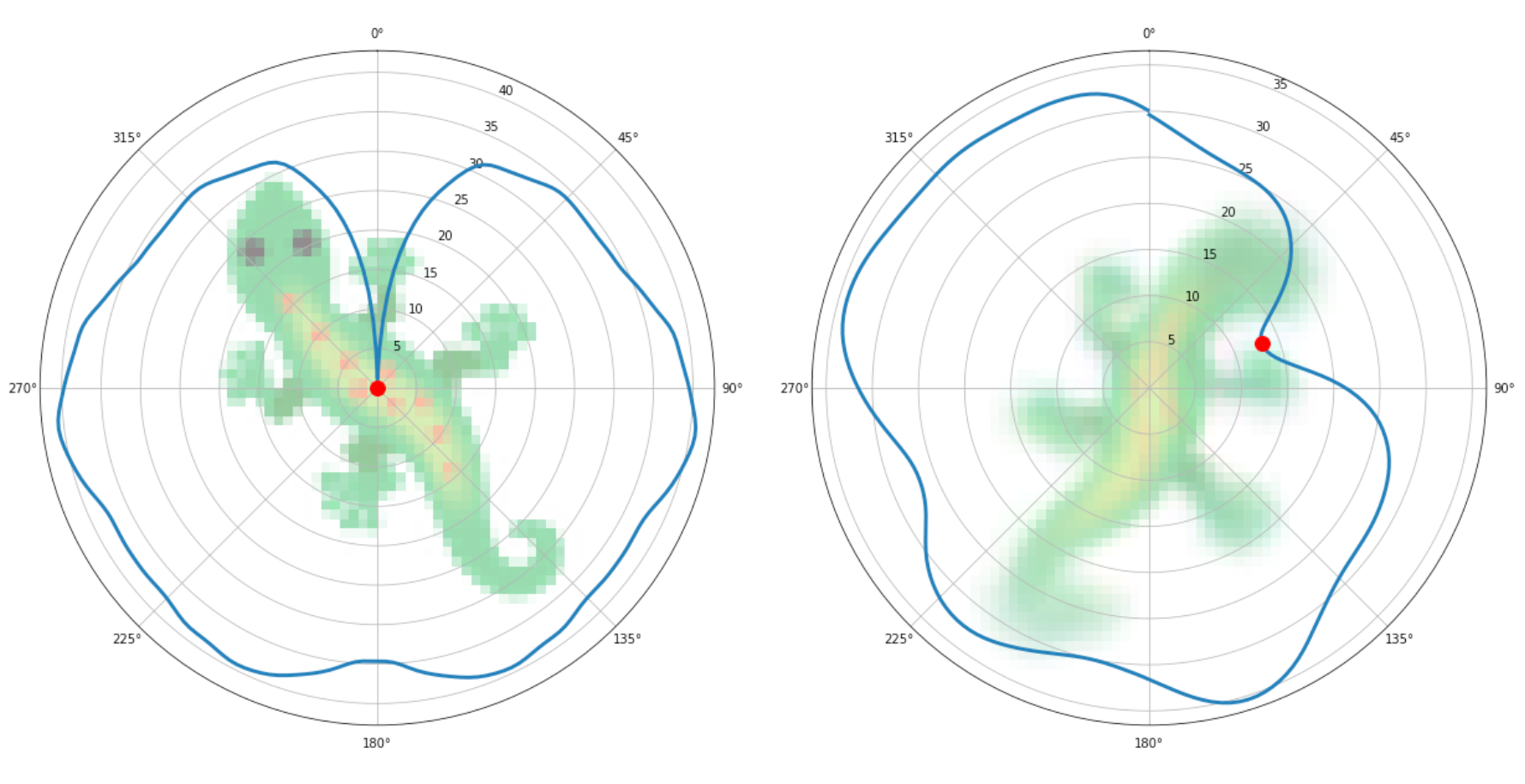}
  \caption{How the rotation-invariant loss gets computed: to the left, as a reference, we calculate the loss with respect to the target pattern for every rotation and see the minimum value is at 0 degrees. To the right, we calculate the loss of a partially-formed pattern during training. The minimum loss can be seen at roughly 60 degrees, and it gets selected for backpropagation.}
  \label{fig:loss_comparison}
\end{figure}

In the single seed strategy, we start from a single seed cell and train Steerable NCA using a loss dependent on the target pattern. Similarly to IsoNCA, we use pixel-wise differences to match the pattern produced by the trained model to the target, but modify the loss function to make it rotation-invariant only and \textit{not} rotation-reflection invariant. Because we cannot know \textit{a priori} what rotation the resulting pattern will have (decided by the random up-down symmetry breaking that could happen along any axis in the 2d plane), we instead select an individual rotation of the target that minimizes the pixel-wise loss value for each NCA-generated sample in the training batch. Note that there is no need for reflection-invariance in the training loss because once one symmetry is broken, the perception scheme of Steerable NCA automatically determines an orthogonal axis always in the same way and, therefore, deterministically breaks left-right symmetry. Figure~\ref{fig:loss_comparison} shows a visual representation of how the minimal loss is selected.

We use the same algorithm presented in IsoNCA to optimize computation, using a polar coordinate transformation and Fast Fourier Transform (FFT) to efficiently compute the discrepancy across different target rotations. We refer to \cite{mordvintsev2022iso} for more implementation details.

\subsubsection{Auxiliary channels}

\begin{figure}[h]
 \centering
  \includegraphics[width=\columnwidth]{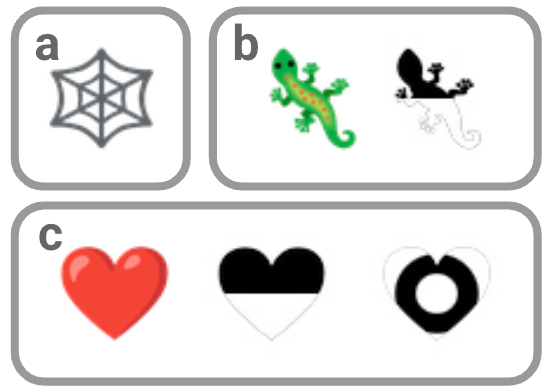}
  \caption{Target patterns and corresponding auxiliary targets used for the rotation-invariant training: \textbf{(a)} the spiderweb does not use auxiliary targets; \textbf{(b)} the lizard uses a binary auxiliary target; \textbf{(c)} the heart uses both binary and radial-encoding auxiliary targets.}
  \label{fig:aux_losses}
\end{figure}

This rotation-invariant loss suffers from the same issues described in IsoNCA, where some patterns exhibit strong local minima within different rotations. We solve this issue in the same way as IsoNCA, using auxiliary channels. Here, we mirror the single-seed experiments in IsoNCA and showcase the same three example patterns: the spiderweb, where no auxiliary loss is needed; the lizard, where a binary auxiliary channel is needed; and the heart, where both the binary and radial-encoding auxiliary channels are needed (see Figure~\ref{fig:aux_losses}).

\section{Results}

\subsection{Two seeds experiments}

\begin{figure}[h]
 \includegraphics[width=\columnwidth]{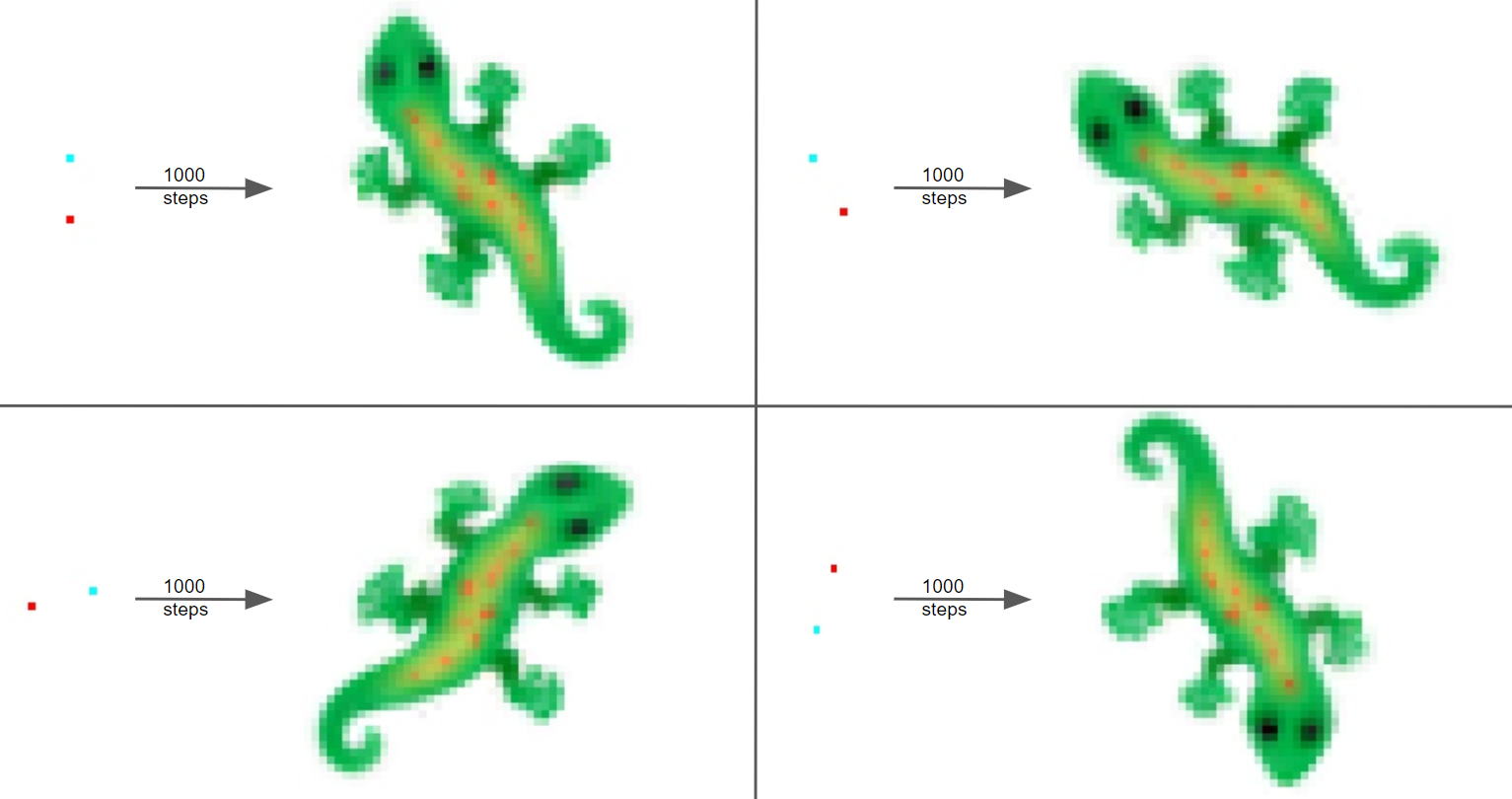}
 \caption{Starting in the top-left corner and moving clockwise, 0 degree, 30 degree, 170 degree, and 280 degree patterns produced by angle-based Steerable NCA trained with pixel-wise L2 loss to grow at 0 degrees. Nonzero degree rotations are inferred by the model during growth based on the arrangement of the starting condition's two constituent seeds.}
 \label{fig:steerable_rotations}
\end{figure}

By adjusting the rotation of our seed structure, we see in Figure~\ref{fig:steerable_rotations} that Steerable NCA are able to self-orient based on their initial conditions. In this figure, we present the angle-based variant, but the gradient-based variant demonstrates identical behavior.

\begin{figure}[h]
 \includegraphics[width=\columnwidth]{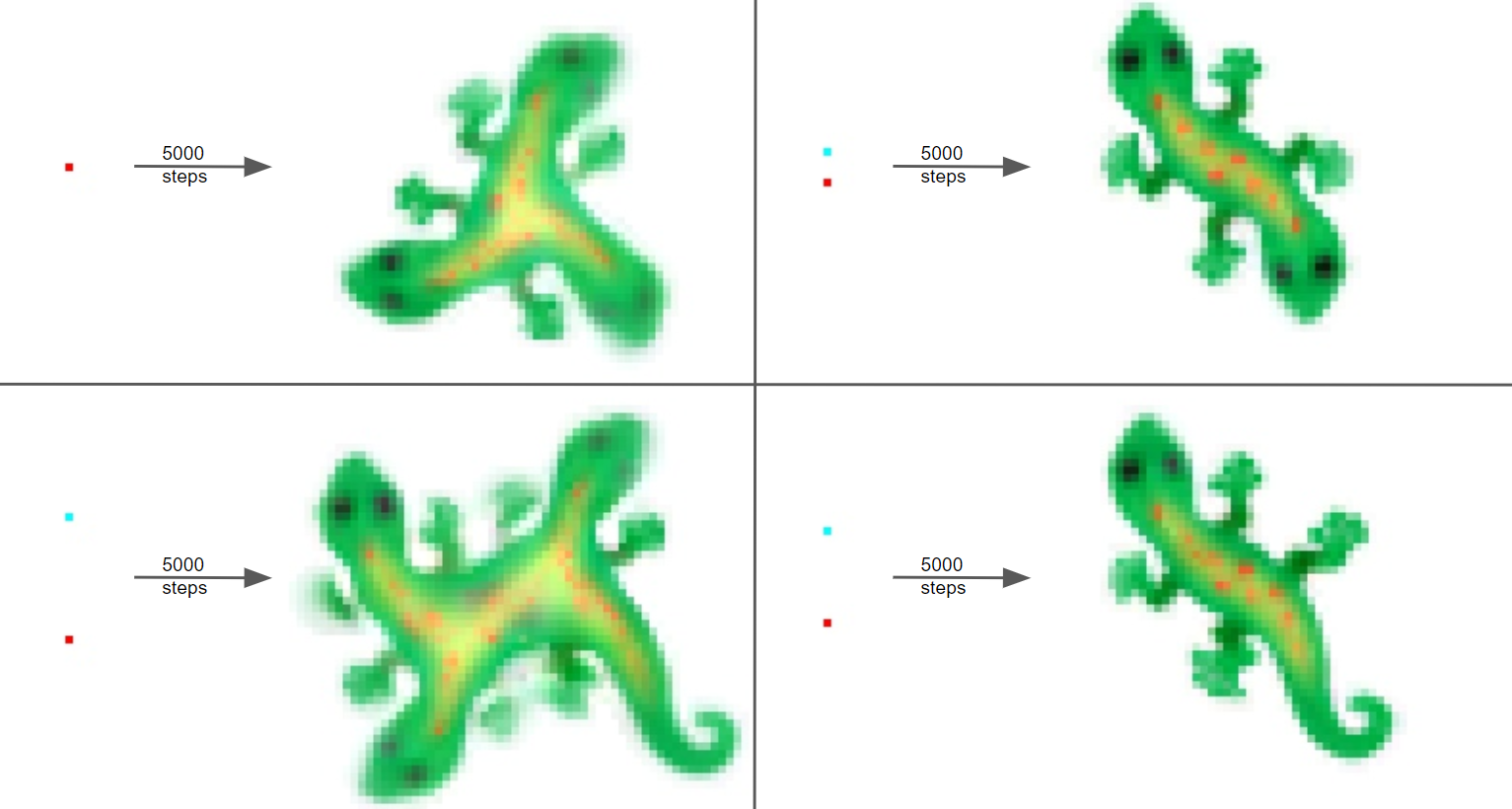}
 \caption{Starting in the top-left corner and moving clockwise, seeds with a diameter of 0 cells, 2 cells, 6 cells, and 8 cells and their resultant patterns produced by angle-based Steerable NCA trained with pixel-wise L2 loss to grow from a seed with a 4 cell diameter at 0 degrees.}
 \label{fig:steerable_mutations}
\end{figure}

Modifying the \textit{diameter} of the seed structure after training further highlights the model's dependence on the intercellular message passing that occurs during the formative stage of model growth. We see this in Figure~\ref{fig:steerable_mutations}, where the cells attempt to align with one another based on the level of separation between the two seeds.

\begin{figure}[h]
 \includegraphics[width=\columnwidth]{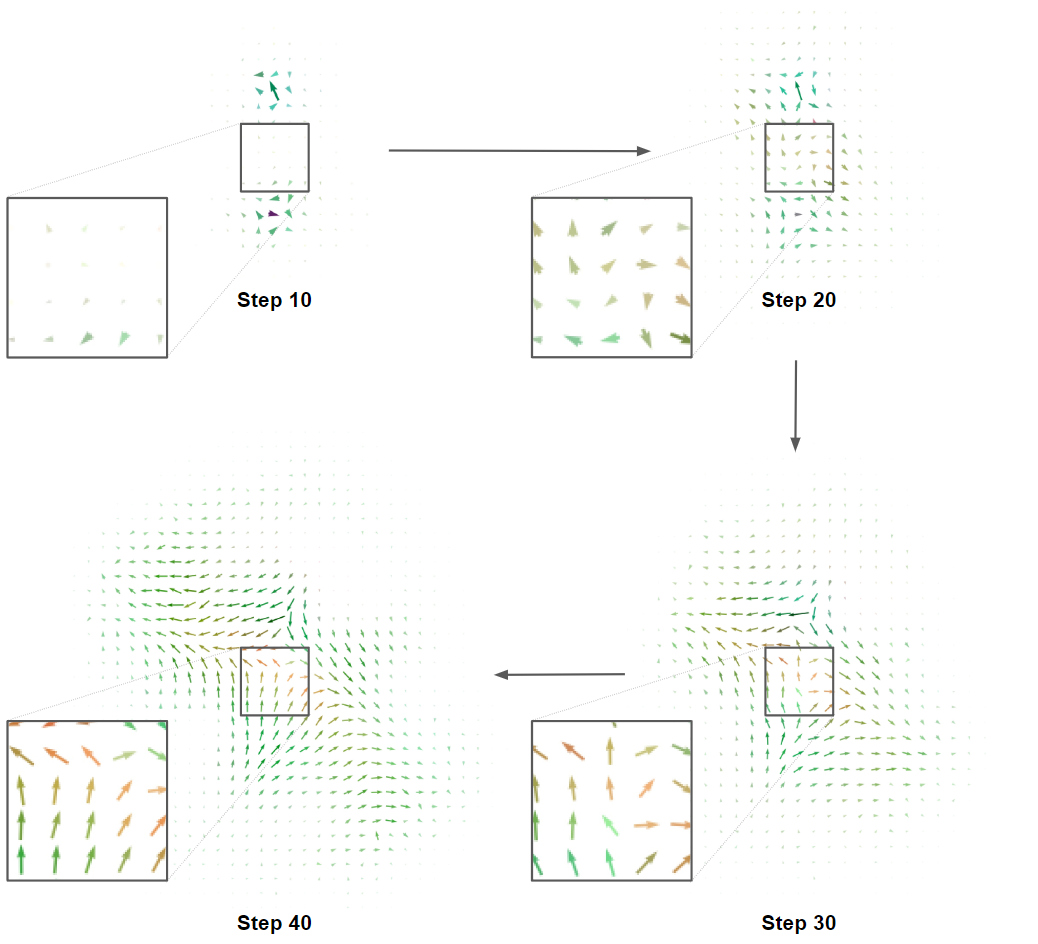}
 \caption{Snapshots of angle-based Steerable NCA angle states during the early alignment phase in which the growths produced by each seed intersect and self-orient.}
 \label{fig:lizard_angle_steps}
\end{figure}

We further analyze this dependence in Figure~\ref{fig:lizard_angle_steps}, wherein we highlight the directional state of angle-based Steerable NCA at snapshots before and after the growths originating from the two seeds intersect. Here, the seed angles are initialized randomly, and the emerging cells progressively self-orient into a coherent vector field that describes the model's internal representation of its own orientation. Moreover, we see that distributing growth between the seeds this way produces an inflection point at which the model "decides" the orientation of the resulting pattern. 

It's worth noting the lack of residue - especially in the later stages of growth - of the two seeds here, in contrast to the single-seed case presented in Figure~\ref{fig:angles_viz}. In the latter scenario, the initial seed position can be discerned even after 5000 steps by observing the vector field formed by the cells' internal angles; in the context of two seeds, however, the original seed configuration is obscured after only 40 steps. This suggests some additional structure necessitated by the rotation-invariant training, dependent on an orientation \textit{selected by} the model rather than one imposed on it. 

\subsection{Single seed experiments}

\begin{figure}[h]
  \includegraphics[width=\columnwidth]{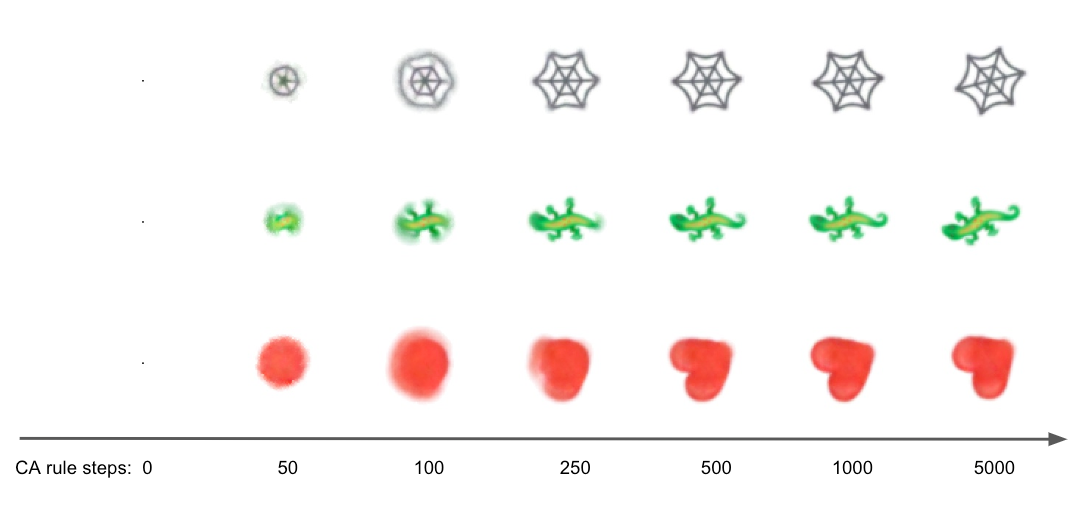}
  \caption{Unfolding of three trained angle-based Steerable NCA rules.}
  \label{fig:single_seed_diff_models}
\end{figure}

\begin{figure}[h]
  \includegraphics[width=\columnwidth]{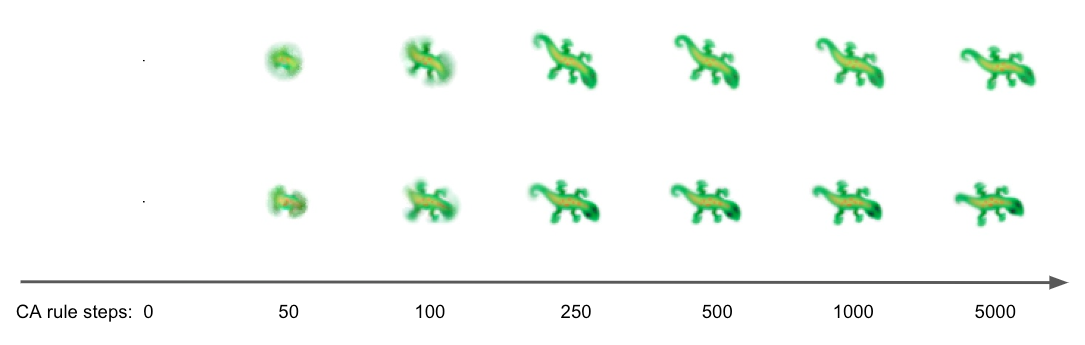}
  \caption{Comparison between angle-based Steerable NCA (top) and gradient-based Steerable NCA (bottom).}
  \label{fig:angle_vs_gradient}
\end{figure}

\begin{figure}[h]
  \includegraphics[width=\columnwidth]{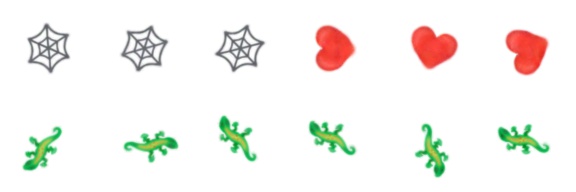}
  \caption{Step 5000 for different runs of three trained angle-based Steerable NCA models.}
  \label{fig:single_seed_different_runs}
\end{figure}

Figure~\ref{fig:single_seed_diff_models} shows the execution in time of three different trained angle-based Steerable NCA models, starting from a single seed and running for up to 5000 steps. Here, we can observe how trained patterns tend to rotate very slowly. This is an artifact of the rotation-invariant loss used, which does not enforce patterns to stand still; every possible rotation of the pattern has the same loss and, thus, a dynamically rotating final state will have the same loss as a static, non-rotating one. Note in Figure~\ref{fig:single_seed_different_runs} how every run of the same model results in a different pattern rotation, but always the same reflection. Figure~\ref{fig:single_seed_diff_models} and Figure~\ref{fig:single_seed_different_runs} show the angle-based model, but gradient-based models manifest similar behaviours. Figure~\ref{fig:angle_vs_gradient} compares the lizard pattern produced by the angle- and gradient-based models.

Training Steerable NCA appears to be faster than training IsoNCA. For instance, training a stable and well-formed lizard takes only 10000 training steps for the Steerable variant, as opposed to 30000 steps for the Isotropic variant with the exact same training configurations. Moreover, the Steerable models empirically grow their target shape faster. This is likely due to Steerable NCA having to break only one symmetry as opposed to the two required by IsoNCA.

\begin{figure}[h]
 \centering
  \includegraphics[width=0.9\columnwidth]{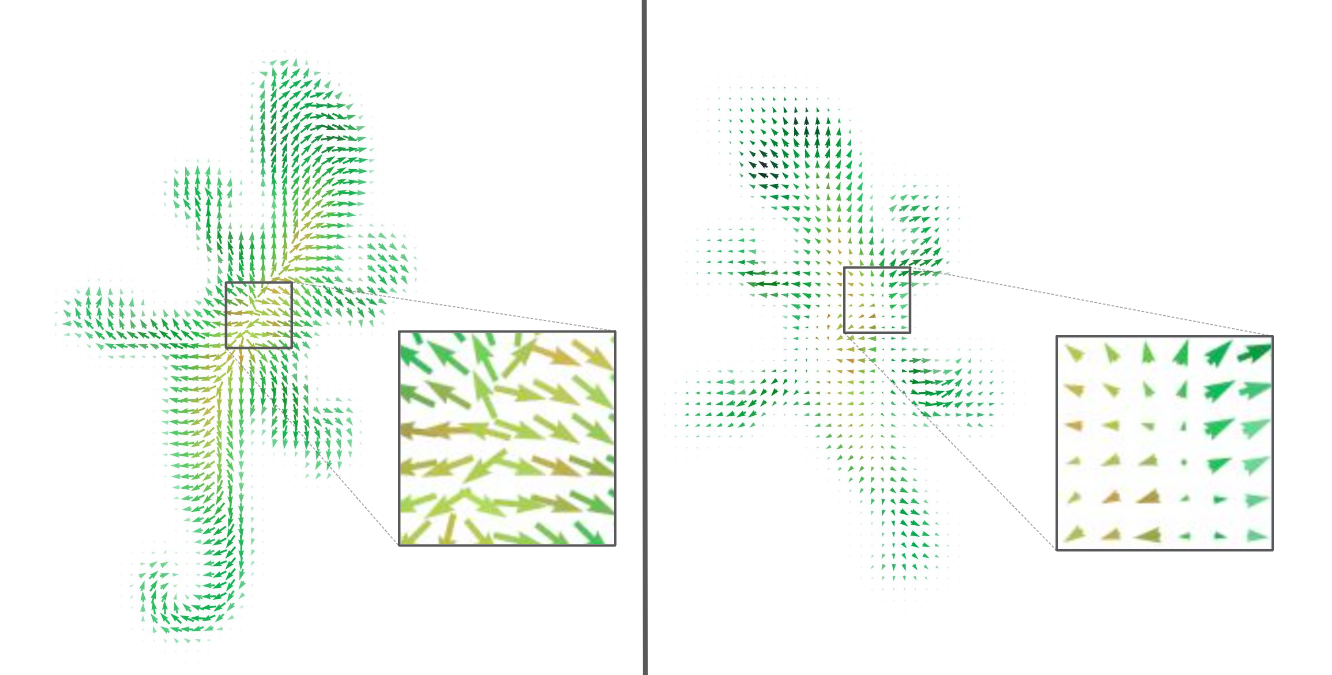}
  \caption{Internal angles for each cell in the Steerable NCA lizard models, snapshot at 5000 steps. Left: an angle-based model in which the angles form a spiral originating at the center. Right: a gradient-based model in which angles cannot have discontinuities, and instead are suppressed, forming a manifold of adirectional cells within the lizard.}
  \label{fig:angles_viz}
\end{figure}

Figure~\ref{fig:angles_viz} shows the internal angles of both Steerable NCA model variants for the lizard pattern, after a time unfolding of 5000 steps. Here, it appears that cells favour having diverging angles from one another. The angles in the angle-based model (left) seem to generate a spiral originating at the center of the pattern field. On the other hand, angles in gradient-based model (right) cannot create discontinuities, as the angle gradients are continuous by construction. However, there appear to be some uniform manifolds where angles are suppressed.

\section{Conclusion and future work}

In this work we demonstrated the capability of chiral NCA models to reliably grow complex asymmetric patterns. We show that they are capable of both inferring a specified orientation based on initial conditions as well as learning to break symmetries independently using a rotation-invariant loss function. Steerable NCA are shown to be simpler to train than their Isotropic counterparts, while maintaining local coordination properties. We believe that this added tool in the differentiable morphogenetic toolbox may be useful in morphogenesis simulations and multi-agent-based modeling, where agents might benefit from having an inner sense of orientation.

\footnotesize
\bibliographystyle{apalike}
\bibliography{example} 

\end{document}